\documentclass{article}

    \PassOptionsToPackage{numbers, compress}{natbib}

    \usepackage[preprint]{neurips_2025}

\usepackage[utf8]{inputenc} 
\usepackage[english]{babel}

\usepackage[T1]{fontenc}    
\usepackage{hyperref}       
\hypersetup{
    colorlinks=true, 
    linkcolor=blue, 
    urlcolor=blue, 
    citecolor=blue 
}
\usepackage{url}            
\usepackage{booktabs}       
\usepackage[table]{xcolor} 
\usepackage{amsfonts}       
\usepackage{nicefrac}       
\usepackage{microtype}      
\usepackage{colortbl}
\usepackage{makecell}
\usepackage{amsmath}    
\usepackage{listings}
\usepackage{graphicx}
\usepackage{siunitx}    
\title{OmniEncoder: See, Hear, and Feel  Continuous Motion Like Humans With One  Encoder}
\author{%
  \textbf{Detao Bai}$^{1}$\thanks{Lead author: detao.bdt@alibaba-inc.com}, 
  \textbf{Shimin Yao}$^{1}$,
  \textbf{Weixuan Chen}$^{1}$,
  \textbf{Chengen Lai}$^{1}$\\ 
  \textbf{Yuanming Li}$^{1}$,
  \textbf{Zhiheng Ma}$^{2}$,
  \textbf{Xihan Wei}$^{1}$\thanks{Project Lead: xihan.wxh@alibaba-inc.com}\\
 \small $^{1}$Tongyi Lab Alibaba Group,$^{2}$ Shenzhen University of Advanced Technology
}
\begin{document}

\maketitle

\begin{abstract}
Recent advances in omni-modal large language models have enabled remarkable progress in joint vision-audio understanding. However, prevailing architectures rely on modality-specific encoders with a \emph{video-coarse, audio-dense} design---sampling visual frames at 1--2 fps while processing audio waveforms at 25 fps---resulting in systems that perceive video \emph{frame by frame, modality by modality} rather than holistically as humans do. Such a discrepancy leaves models with impoverished cross-modal interaction during encoding and an inability to capture fine-grained visual motion. To bridge this gap, we present \textbf{Omni-Encoder, a unified Transformer backbone designed to co-embed visual and audio signals at a symmetrical 25 fps} within a shared latent space. This architecture leverages three core innovations---the Omni-Encoder Token Template, Omni-RoPE, and Temporal Window Shifting---to effectively reconcile the dual challenges of modality disentanglement and computational efficiency. Experiments demonstrate that, compared to the modality-specific baseline Qwen2.5-Omni under the same input token budget to the LLM decoder, Omni-Encoder delivers substantial gains on visual continuous understanding tasks---such as sign language recognition  and fine-grained sports action analysis---while maintaining competitive performance on established audio-visual benchmarks such as AVQA and Speaker Identification and Localization. These results suggest that unified omnivorous encoding offers a promising direction for building omni-modal models that more closely reflect the integrated nature of human perception.

\end{abstract}

\section{Introduction}

\begin{figure}[ht]
\centering
\includegraphics[width=0.8 \linewidth]{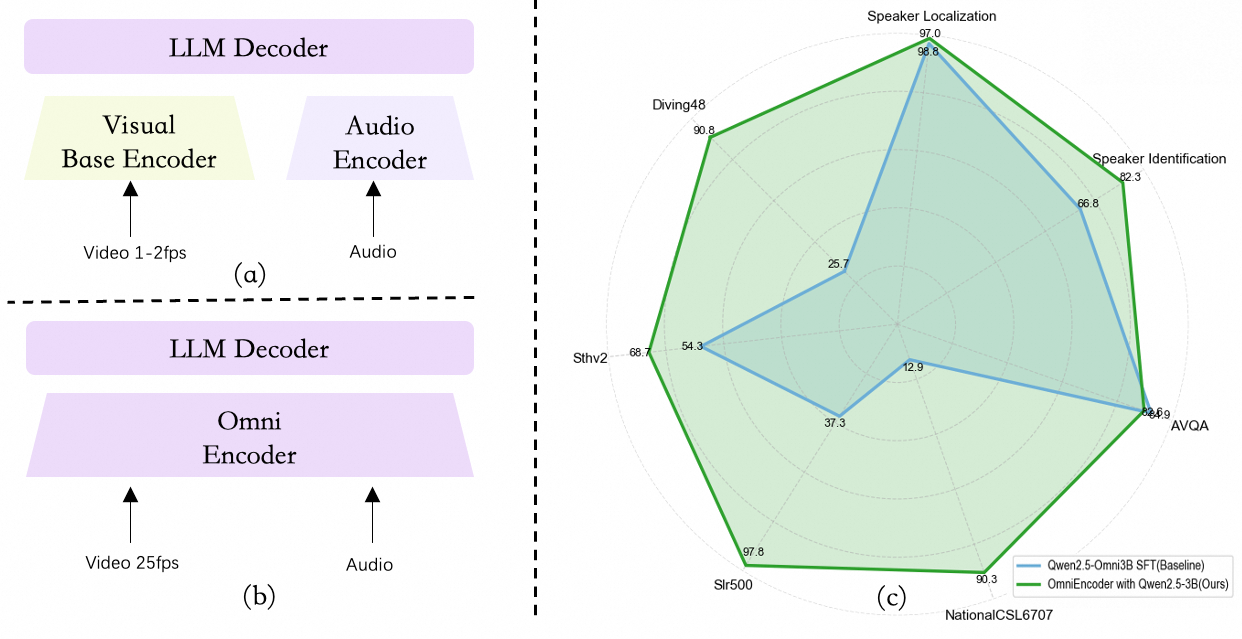}
\caption{\label{fig:oldomni}\textbf{ (a). Modality-specific encoders: }process visual and audio information separately at mismatched frame rates (1--2 fps for video, 25 fps for audio).\textbf{(b). Omni-Encoder:} a single Transformer jointly encodes audio, visual base, and visual continuous tokens at 25 fps within a unified representation space.\textbf{(c).} With the same number of input tokens to the LLM, we compare Qwen2.5-Omni-3B with the original encoder and Omni-Encoder, both trained with SFT, on tasks requiring visual continuous understanding (e.g. sign language, sports action) and audio-visual QA tasks.}
\vspace{-0.3cm}
\end{figure}

Human-beings understand the world is inherently omnimodal, involving the integration of visual, auditory, and linguistic information. Benefiting from the rapid advances in vision~\cite{clip,siglip,siglip2}, audio~\cite{whisper,qwen-audio,qwen2-audio}, and language~\cite{qwen3.5,LLaMA} models, omni-modal LLMs~\cite{Qwen3.5-Omni,Qwen3Omni,Qwen25Omni,Gemini3,Ola,VITA-1.5,HumanOmni,HumanOmniSpeaker} have grown remarkably capable at vision-audio understanding.

Most OmniLLMs typically follow a modality-specific modular design. Specifically, a visual encoder~\cite{siglip,siglip2}, an audio encoder~\cite{whisper,qwen2-audio}, and modality-specific projectors are utilized to encode each modality's information separately (Fig.~\ref{fig:oldomni}a). While this design philosophy effectively leverages modality-specific pretraining, such technological inertia results in omni-models that perceive video ``frame by frame, modality by modality''. Yet this does not reflect how humans perceive video. In contrast, humans see, hear, and feel continuous motion simultaneously. Cross-modal interactions occur at the earliest stages of sensory processing in the central nervous system~\cite{meredith1987determinants,mishra2007early}. These observations naturally motivate a fundamental question: \emph{Is it possible to jointly encode visual and audio information within a unified omnivorous encoder?}

Achieving this presents two key challenges. First, processing multiple modalities within a unified encoder requires the model to distinguish modality-specific representations, support diverse encoding modes (visual-only, audio-only, and visual-audio), and retain the strong extrapolation capabilities of prior modality-specific encoders (e.g., support for arbitrary resolutions and temporal lengths). Second, constrained by the computational cost of bidirectional attention, existing OmniLLM works adopt a \emph{video-coarse, audio-dense} design --- for instance, in Gemini and Qwen2.5-Omni, video is sampled at 1--2 fps while audio is processed at 25 fps. For videos with dense motion dynamics (e.g., sign language recognition, fine-grained sports such as gymnastics, or lip reading),it is desirable for the model to capture fine-grained motion at a frame rate comparable to audio, while maintaining acceptable computational overhead.

To this end, we introduce Omni-Encoder (Fig.~\ref{fig:oldomni}b), a single Transformer backbone that jointly encodes visual frames and audio waveforms within one representation space at high frame rate (25fps). The key designs of our Omni-Encoder are threefold: 

\begin{itemize}
\item \textbf{Omni-Encoder Token Template.} We propose a unified omni-modal encoding template that decomposes raw 25\,fps video into three token streams---Audio, Visual Continuous (VC), and Visual Base (VB) tokens---at each temporal position. VC tokens are introduced as frame-wise learnable queries that capture inter-frame motion dynamics, such as motion trajectories, gesture onsets, and micro-movements. The encoder output passes through a Token Sparsifier that reduces VB tokens to 2\,fps while preserving Audio and VC tokens at full 25\,fps, drastically cutting the token count forwarded to the downstream decoder without sacrificing motion fidelity.
    \item \textbf{Omni-RoPE.} A 3D rotary positional encoding that assigns each token a unique coordinate $(t, h', w')$ in a unified spatio-temporal-modality space. This formulation enables the Omni-Encoder to distinguish heterogeneous Audio, Visual Continuous, and Visual Base tokens and model their cross-modal relationships, while preserving resolution-agnostic extrapolation for arbitrary visual resolutions and temporal lengths.
    \item \textbf{Temporal Window Shifting.} We propose a joint spatiotemporal attention mechanism that alternates between local and shifted temporal windows across layers, reducing complexity from $\mathcal{O}(T^2)$ to $\mathcal{O}(T \cdot G)$. This enables dense cross-modal interaction at 25 fps while keeping computation linear in video length.
\end{itemize}

\section{Architecture}

\begin{figure}[ht]
\vspace{-0.3cm}
\centering
\includegraphics[width=0.96 \linewidth]{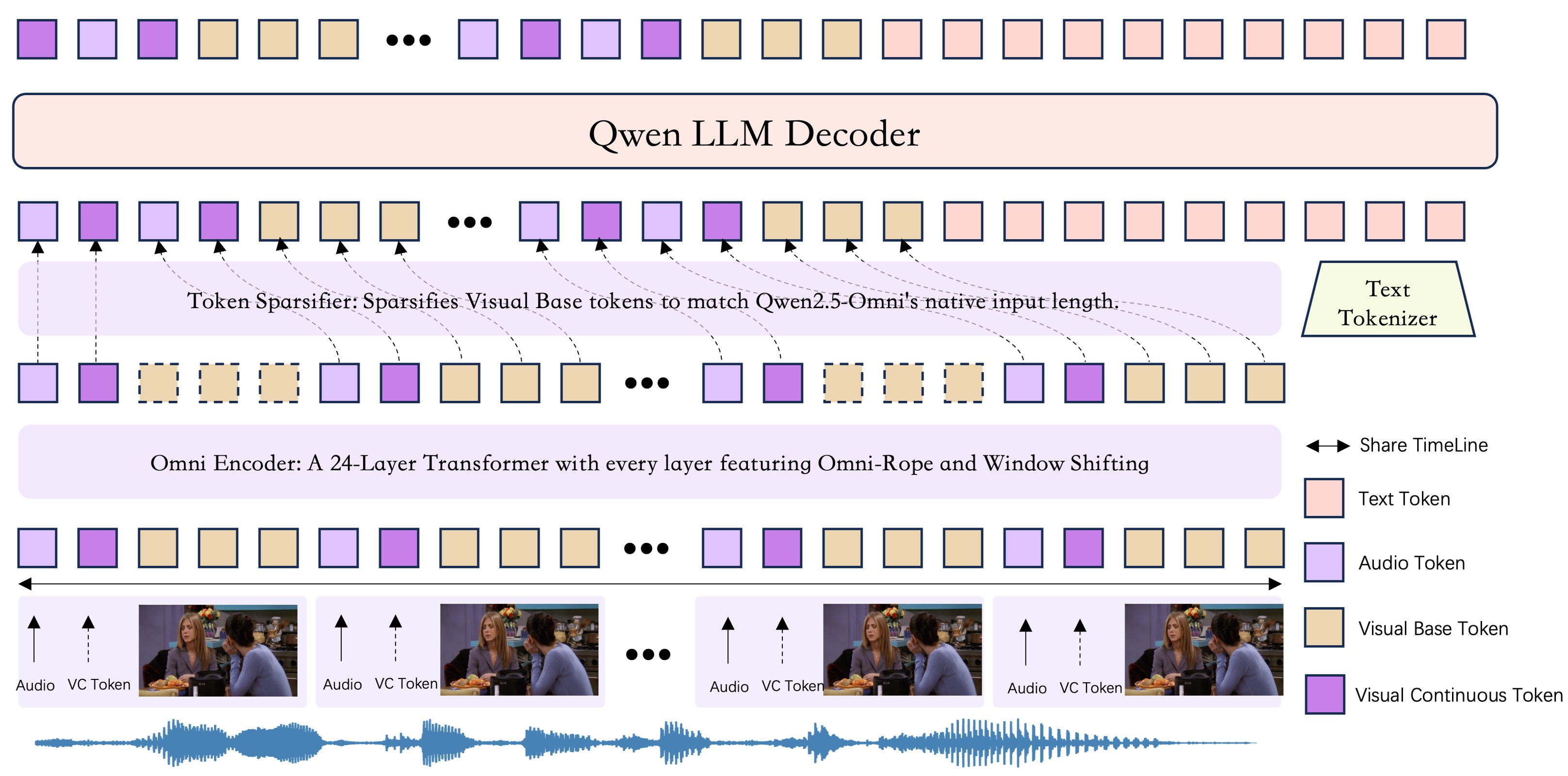}
\caption{\label{fig:archmodel} \textbf{Architecture of Omni-Encoder.} A 24-layer Transformer jointly encodes Audio, Visual Continuous, and Visual Base tokens from raw 25\,fps video through unified self-attention, with each layer incorporating Omni-RoPE and Temporal  Window Shifting. A Token Sparsifier then sparsifies Visual Base tokens to match the native input length of Qwen2.5-Omni.}
\vspace{-0.3cm}
\end{figure}

As illustrated in Figure~\ref{fig:archmodel}, Omni-Encoder is a 24-layer Transformer backbone that natively processes visual frames, audio waveforms, and continuous motion signals at the original video frame rate of 25 fps. Rather than encoding each modality in isolation and fusing them at a later stage, our approach co-embeds all modalities into a unified sequence and applies joint self-attention from the first encoder layer, allowing cross-modal alignment to emerge during encoding rather than as a downstream patch. This design mirrors how humans integrate sight, sound, and motion to interpret social interactions---enabling holistic, temporal  perception from raw video input.

The encoded token sequence passes through a Token Sparsifier before entering the Qwen-LLM 3B decoder for downstream reasoning. Critically, the Token Sparsifier strategically retains all Audio and Visual Continuous tokens at full temporal resolution while sparsely sampling Visual Base tokens, reducing the total token count to match the native input length of Qwen2.5-Omni~\cite{Qwen25Omni}. This ensures full compatibility with the pretrained decoder without any architectural modification, while preserving high temporal resolution for capturing fine-grained behavioral cues.

To enable efficient processing of native 25 fps video, Omni-Encoder incorporates three key architectural designs: Omni-Encoder Token Template, Omni-RoPE, and Temporal Window Shifting. We describe each component in the following subsections.

\subsection{Omni-Encoder Token Template:Jointly Encoded Audio, VC \& VB Tokens}
\label{app:TokenTemplate}

As shown in Figure~\ref{fig:archmodel}, input video is decomposed into three distinct token streams:
\begin{itemize}
    \item \textbf{Audio Tokens($\mathbf{a}_t$):} Audio tokens are extracted from audio log-Mel spectrograms using two lightweight 1D CNNs, producing discrete audio embeddings aligned with video frames.
    \item \textbf{Visual Continuous Tokens($\mathbf{v}_t^{\mathrm{c}}$ ):} VC tokens are introduced as frame-wise learnable vectors at each temporal position, capturing inter-frame information during encoding such as motion cues, gaze direction, and other high-frequency dynamics.
    \item \textbf{Visual Base Tokens($\mathbf{v}_t^{\mathrm{b}}$ ):} VB tokens are generated via patch embedding, which partitions each frame into non-overlapping spatial patches and projects them into a unified embedding space, capturing static appearance features such as texture, object shape, and scene layout.

\end{itemize}

As defined in Eq.~\eqref{eq:token_in}, the input token sequence $\mathbf{T}_{\text{in}}$ concatenates three modalities at each temporal position: one Audio token $\mathbf{a}_t$, one Visual Continuous token $\mathbf{v}_t^{\mathrm{c}}$, and a set of Visual Base tokens $\mathbf{v}_t^{\mathrm{b}}$ from spatial patch embedding, where the number of Visual Base tokens per frame scales with input resolution.The Token Sparsifier $\mathcal{S}(t)$ (Eq.~\eqref{eq:delta_r}) transforms $\mathbf{T}_{\text{in}}$ into $\mathbf{T}_{\text{out}}$ (Eq.~\eqref{eq:token_out}) by selectively preserving Visual Base tokens: $\mathcal{S}(t) = 1$ retains them at frame $t$, while $\mathcal{S}(t) = 0$ drops them entirely. The resulting $\mathbf{T}_{\text{out}}$ reduces the total token count to match the native input length of Qwen2.5-Omni~\cite{Qwen25Omni}, ensuring compatibility with the pretrained decoder while preserving high-fidelity motion and audio encoding.

\begin{subequations}\label{eq:token_formulation}
\begin{align}
\mathbf{T}_{\text{in}} &= \bigoplus_{t=1}^{T\cdot f_v} \left( \mathbf{a}_t \oplus \mathbf{v}_t^{\mathrm{c}} \oplus  \mathbf{v}_{t}^{\mathrm{b}} \right) \label{eq:token_in} \\
\mathbf{T}_{\text{out}} &= \bigoplus_{t=1}^{T\cdot f_v} \left( \mathbf{a}_t \oplus \mathbf{v}_t^{\mathrm{c}} \oplus \mathcal{S}(t) \cdot \mathbf{v}_{t}^{\mathrm{b}} \right) \label{eq:token_out} \\
\mathcal{S}(t)  &= \begin{cases} 
1, & \text{if } t \bmod (f_v / f_v^b) = 0  \\ 
0, & \text{otherwise} 
\end{cases} \label{eq:delta_r}
\end{align}
\end{subequations}
where $f_v = 25\,\text{fps}$ is the input video frame rate, $T$ is the video duration in seconds, and $f_v^{\mathrm{b}} = 2\,\text{fps}$ is the target frame rate of Visual Base tokens after downsampling.

To further reduce computational overhead during training, we apply Tubelet embedding~\cite{ViViT} to both $\mathbf{T}_{\text{in}}$ and $\mathbf{T}_{\text{out}}$:
\begin{align}
\mathbf{T}'_{\text{in}} = \mathcal{D}_{\tau}(\mathbf{T}_{\text{in}}), \qquad 
\mathbf{T}'_{\text{out}} = \mathcal{D}_{\tau}(\mathbf{T}_{\text{out}})
\end{align}
where $\mathcal{D}_{\tau}(\cdot)$ denotes the tubelet downsampling operator with a temporal tubelet size of $\tau = 2$, aggregating every $\tau$ consecutive frames into a single token. This reduces the total token count processed during encoding by half.

\subsection{Omni-RoPE: 3D Rotary Positional Encoding for Multimodal Tokens}

Rotary Position Embedding (RoPE)~\cite{Rope} encodes relative positional information through rotation matrices and has become the standard positional encoding in modern large language models. In video large language models, 3D RoPE extends the positional index to a $(t, h, w)$ triplet, corresponding to temporal, height, and width dimensions, respectively.

Directly applying 3D RoPE to our three modalities introduces a fundamental conflict: the $(h, w)$ spatial plane is fully occupied by VB tokens starting from the origin, leaving Audio and VC tokens without independent coordinates and rendering them indistinguishable to the attention mechanism.

\begin{figure}[ht]
\centering
\includegraphics[width=0.96 \linewidth]{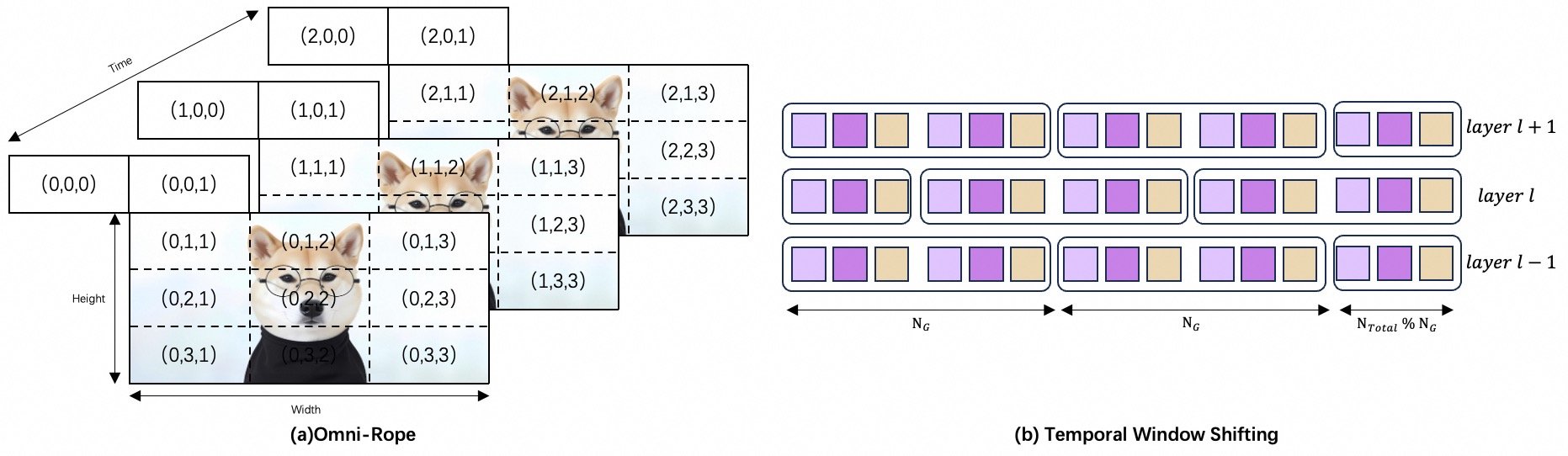}
\caption{\label{fig:omnirope} Omni-Rope and Temporal Window Shifting .\textbf{(a)Omni-Rope:} 3D rotary encoding assigns unique $(t, h', w')$ coordinates—Audio at $(t,0,0)$, Visual Continuous at $(t,0,1)$, Visual Base starting from $(t,1,1)$. 
\textbf{(b) Temporal Window Shifting:} Alternating local window and shifted window attention reduces complexity from $\mathcal{O}(T^2)$ to $\mathcal{O}(T \cdot G)$. $G$: group size (frames per window); $N_G$: tokens per group; $N_{\text{Total}}$: total tokens.}
\vspace{-0.3cm}
\end{figure}

As show in  Figure~\ref{fig:omnirope}(b), We resolve this by shifting VB token spatial coordinates by $(+1, +1)$, formally remapping the original $(t, h, w)$ to $(t, h', w')$, where:

\begin{equation}
(h', w') = (h + 1,\, w + 1).
\end{equation}

The vacated origin is then partitioned between Audio tokens at $(t, 0, 0)$ and VC tokens at $(t, 0, 1)$. Under this remapping, the rotation angle for a token at position $(t, h', w')$ becomes:
\begin{equation}
\theta_k(t, h', w') = \frac{t \cdot \omega_{k,t} + h' \cdot \omega_{k,h} + w' \cdot \omega_{k,w}}{10000^{2k/d}}
\label{eq:omni_rope}
\end{equation}
where $(h', w')$ for VB tokens, and $(0, 0)$ or $(0, 1)$ for Audio and VC tokens, respectively. This assignment ensures all three modalities receive non-overlapping positional encodings within the same 3D RoPE formulation, without any modification to its mathematical definition or implementation.

\subsection{Temporal Window Shifting: Efficient Joint Spatiotemporal Attention}

Popular audio encoders use full temporal attention for speech continuity\cite{whisper}, while visual encoders employ either intra-frame spatial attention\cite{clip,siglip} or GOP-based spatiotemporal attention\cite{VideoMAE,V-JEPA2}. The Omni-Encoder must process both modalities simultaneously, but global spatiotemporal attention across all frames incurs quadratic complexity $\mathcal{O}(T^2)$ with  respect to sequence length, rendering native 25,fps video infeasible on current hardware.

To address this challenge, we propose Temporal Window Shifting---a structured attention mechanism inspired by Swin Transformer~\cite{SwinTransformer}, which alternates between GOP-based spatiotemporal window attention and shifted  temporal window across Transformer layers (Fig.~\ref{fig:omnirope}(b)).The token sequence is partitioned into non-overlapping temporal groups of $G=16$ frames, and joint spatiotemporal attention is computed independently within each group. In alternating layers, the window boundaries are shifted by $G/2$ frames, so that each new group overlaps with two adjacent groups from the previous layer, enabling cross-group information flow. Since attention is confined to groups of $G$ frames, the complexity reduces from $\mathcal{O}(T^2)$ to $\mathcal{O}(T \cdot G)$. With $G \ll T$, the computational cost grows linearly with video length.

Formally, let $N_G$ denote the number of tokens per group. At layer $l$, attention is computed within each group $i = 1, \dots, \lceil T/G \rceil$:
\begin{align}
\text{GroupAttention}^{(l)}_i &= \text{Softmax}\left(\frac{\mathbf{Q}_i \mathbf{K}_i^\top}{\sqrt{d}}\right)\mathbf{V}_i, \quad \mathbf{Q}_i, \mathbf{K}_i, \mathbf{V}_i \in \mathbb{R}^{N_G \times d}.
\end{align}
At layer $l+1$, the window is shifted by $G/2$ frames,$N_{\text{frame}}$  denote the number of tokens per frame:
\begin{align}
\text{ShiftedGroup}^{(l+1)}_j &= \left\{ \mathbf{t}_k \mid (j \cdot G - G/2) \cdot N_{\text{frame}} \leq k < ((j+1) \cdot G - G/2) \cdot N_{\text{frame}} \right\}
\end{align}

\section{Training}

In terms of training methodology, we adopt a \textbf{single-stage, end-to-end training strategy}. We do not pre-train the Omni-Encoder separately, nor do we introduce an additional projection layer to map its output into the language model’s input space. Instead, we perform joint training directly on top of the LLM.

Specifically, we employ the LLM component of\textbf{ Qwen2.5-Omni-3B}\cite{Qwen25Omni} as our generation decoder.This LLM has already been pre-trained on multimodal data and thus possesses strong cross-modal understanding capabilities, enabling it to effectively process the unified representations of vision, audio, and text. During training,\textbf{only final language modeling (LM) head} are trainable; all other LLM parameters are frozen.In this setup, the entire LLM functions as a powerful, fixed-parameter “big decoder head.”
This training strategy avoids complex multi-stage pre-training pipelines, ensuring strong model expressiveness while significantly improving training efficiency and stability.

To facilitate training and accelerate convergence, we initialize the Omni-Encoder with weights pre-trained via self-supervised video representation learning~\cite{V-JEPA2}, while all newly introduced components—including the audio embedding layers and learnable queries for Visual Continuous tokens—are randomly initialized and trained from scratch.

\section{Experiment}

Table~\ref{tab:benchmarks} presents the results of Omni-Encoder across visual continuous understanding and audio-visual reasoning benchmarks, covering motion and action recognition (Diving48~\cite{diving48}, SthSv2~\cite{sthv2}), sign language recognition (SLR500~\cite{islr500}, NationalCSL6707~\cite{NationalCSL6707}), audio-visual question answering (AVQA~\cite{AVQA}), and speaker localization and identification~\cite{HumanOmniSpeaker}. We further evaluate Omni-Encoder on audio-visual  recognition (VSR, ASR, AVSR) using the LRS2 dataset~\cite{lrs2}. Based on the experimental results, we make four key observations. 

\noindent\textbf{Observation 1: Current Omni models lack frame-level motion modeling for continuous video understanding.}
State-of-the-art closed-source Omni models such as Gemini-2.5-Pro~\cite{Gemini25} and Qwen3.5-Omni~\cite{Qwen3.5-Omni} perform at near-random levels on video-level continuous understanding tasks, achieving only 4.94\% and 6.86\% on Diving48\cite{diving48}, and 1.2\% and 0.8\% on SLR500\cite{islr500}, respectively. This performance gap reflects the sparse visual encoding strategy adopted by existing OmniLLMs---where video frames are sampled at 1--2 fps and processed independently---which is insufficient for capturing the dense, frame-by-frame motion dynamics required for fine-grained action and gesture recognition. Notably, this limitation persists even under supervised fine-tuning: the SFT variant of Qwen2.5-Omni-3B\cite{Qwen25Omni} reaches only 25.7\% on Diving48 and 37.3\% on SLR500, despite being trained directly on task-specific labels. The observation that task-specific supervision yields only modest improvements suggests that the bottleneck stems from the coarse-grained, frame-independent visual encoding, rather than from insufficient task adaptation alone.


\begin{table}[htbp]
\centering
\small 
\vspace{0.2cm}
\caption{Omni-Encoder performance across visual continuous understanding and audio-visual reasoning benchmarks, compared with domain-specific models and Omni models. Arrows ($\uparrow$) indicate higher is better.}
\label{tab:benchmarks}
\resizebox{0.96\textwidth}{!}{
\begin{tabular}{cccccccc}
\toprule
\textbf{Method} & \multicolumn{2}{c}{\textbf{Motion \& Action}} & \multicolumn{2}{c}{\textbf{Sign Language}} & \multicolumn{1}{c}{\textbf{AVQA}} & \multicolumn{2}{c}{\textbf{Speaker}} \\
 & \textbf{Diving48}($\uparrow$) & \textbf{Sthv2}($\uparrow$) & \textbf{SLR500}($\uparrow$) & \textbf{NationalCSL6707}($\uparrow$) & \textbf{AVQA}($\uparrow$) & \textbf{Localization}($\uparrow$) & \textbf{Identification}($\uparrow$) \\ 
\midrule
\rowcolor{gray!20} \multicolumn{8}{c}{Closed-source Omni Models} \\
\addlinespace
 Gemini-2.5-pro~\cite{Gemini3} & 4.94 & - & 1.2 & - & - & - & - \\
 Qwen3.5-Omni~\cite{Qwen3.5-Omni} & 6.86 & - & 0.8 & - & - & - & - \\
\midrule
\rowcolor{gray!20} \multicolumn{8}{c}{Specific Models without LLM} \\
\addlinespace
 Signbert~\cite{signbert} & - & - & 97.6 & - & - & - & - \\
 NationalCSL-DP~\cite{NationalCSL6707} & - & - & - & 69.61 & - & - & - \\
 Cat~\cite{Cat} & - & - & - & - & 92.0 & - & - \\
 SigLIP2~\cite{siglip2} & 75.3 & 49.9 & - & - & - & - & - \\
 InternVideo2s2-1B~\cite{InternVideo2} & 86.4 & 69.7 & - & - & - & - & - \\
 V-JEPA2-L~\cite{V-JEPA2} & 86.0 & 73.7 & - & - & - & - & - \\
\midrule
\rowcolor{gray!20} \multicolumn{8}{c}{MLLMs Processing Dense Video} \\
\addlinespace
 OV-Encoder Codec~\cite{onevisionencoder} & 69.4 & 60.1 & - & - & - & - & - \\
 F-16~\cite{F-16} & 86.5 & - & - & - & - & - & - \\
 VL-JEPA~\cite{VL-JEPA} & 90.1 & 73.2 & - & - & - & - & - \\
 HumanOmni-Speaker~\cite{HumanOmniSpeaker} & - & - & - & - & - & 99.4 & 78.9 \\
\midrule
 Sota Before & 90.3 & 75.3 & 97.6 & 69.61 & 92.0 & 99.4 & 78.9 \\
 Qwen2.5-Omni3B SFT~\cite{Qwen25Omni} & 25.7 & 54.3 & 37.3 & 12.9 & 84.9 & 97.0 & 66.8 \\
\midrule
\rowcolor{gray!20} \textbf{Omni-Encoder} & \textbf{90.8} & \textbf{68.7} & \textbf{97.8} & \textbf{90.32} & \textbf{82.6} & \textbf{98.8} & \textbf{82.31} \\
\bottomrule
\end{tabular}
}
\end{table}

\noindent\textbf{Observation 2: Omni-Encoder matches or exceeds specialists on high-density visual tasks.}
Tasks that demand fine-grained modeling of dense temporal patterns represent the most challenging regime for unified encoders. Unlike prior OmniLLMs that rely on sparse 1--2 fps sampling, Omni-Encoder processes video at 25 fps through frame-wise VC tokens that explicitly capture inter-frame motion dynamics. On the continuous sign language recognition benchmark NationalCSL~\cite{NationalCSL6707}, a large-scale dataset comprising 6,707 distinct sign language glosses, Omni-Encoder achieves 90.32\%, surpassing the previous best domain-specific model (NationalCSL-DP~\cite{NationalCSL6707}, 69.61\%) by over 20 percentage points---a substantial margin that demonstrates dense frame-level motion modeling is particularly effective for capturing the fine-grained gesture distinctions and continuous motion trajectories characterizing sign language. This advantage extends to other benchmarks: on SLR500, our model reaches 97.8\%, outperforming the specialist SignBERT~\cite{signbert}; on Diving48, it achieves 90.8\%, exceeding the previous best of 90.3\%; and on SthSv2, which requires fine-grained reasoning about object--action interactions, our model reaches 68.7\%, competitive with strong baselines such as InternVideo2-S2-1B\cite{InternVideo2} (69.7\%). Collectively, these results establish that a single unified encoder can achieve specialist-level performance across both sign language and action understanding without task-specific architectural customization, validating high-density temporal modeling as an effective general-purpose representation strategy.

\noindent\textbf{Observation 3: Omni-Encoder achieves competitive audio-visual reasoning with a unified architecture.}
On cross-modal reasoning tasks that require joint processing of audio and visual signals, Omni-Encoder performs on par with dedicated dual-tower architectures. On speaker localization and identification tasks, Omni-Encoder matches or surpasses HumanOmni-Speaker~\cite{HumanOmniSpeaker}: it achieves 98.8\% on localization (vs. 99.4\%) and 82.31\% on identification (vs. 78.9\%).On the AVQA benchmark, Omni-Encoder achieves 82.6\%, matching the SFT result of Qwen2.5-Omni-3B~\cite{Qwen25Omni}. Omni-Encoder, by contrast, handles all modalities through a single unified encoder without task-specific customization. Taken together, these findings suggest that unified encoding does not fundamentally limit multi-modal reasoning, and that Omni-Encoder offers a reasonable balance between task generality and performance.


\noindent\textbf{Observation 4: Omni-Encoder demonstrates effective audio-visual recognition without modality-specific preprocessing.}
To further assess the multi-modal capabilities of Omni-Encoder in a recognition setting, we evaluate it on the LRS2 dataset~\cite{lrs2} across VSR, ASR, and AVSR tasks. In the visual-only setting, without any additional preprocessing such as lip cropping, the VSR task achieves a WER of 45.3\%, surpassing the CTC/Attention baseline~\cite{ctc_attentin} and demonstrating the model's ability to effectively extract visual phoneme features from raw video frames. In the audio-only setting, the ASR task yields a WER of 10.2\%, validating the model's ability to efficiently encode audio signals for high-precision speech recognition. When both modalities are available, the AVSR task further reduces the WER to 7.2\%, a 3.0 percentage point improvement over audio alone, demonstrating that the visual stream provides substantial complementary information and that Omni-Encoder effectively fuses cross-modal signals. It is worth noting that these results still lag behind state-of-the-art proprietary models to some extent. We attribute this primarily to the limited training data scale---fewer than 1,000 hours of speech were used for the task. Nevertheless, these experiments sufficiently demonstrate the strong multi-modal modeling potential of Omni-Encoder in audio-visual recognition scenarios.

\section{Conclusion}

In this work, we presented Omni-Encoder, a unified Transformer that jointly encodes visual and audio signals at 25 fps within a single representation space. Through three key innovations---the Omni-Encoder Token Template, Omni-RoPE, and Temporal Window Shifting---our approach enables frame-level synchronized audio-visual modeling at high temporal density while preserving computational efficiency. Experiments show that Omni-Encoder achieves state-of-the-art or competitive performance across both fine-grained visual understanding and audio-visual reasoning tasks, without increasing the token budget forwarded to the downstream LLM. These results demonstrate that unified omnivorous encoding offers a promising direction for omni-modal models that better reflect the integrated nature of human perception.

{
\medskip
\small  
\bibliographystyle{unsrt}
\bibliography{main}
}


\end{document}